\pdfoutput=1

\documentclass[11pt]{article}

\usepackage[review]{acl}

\usepackage{times}
\usepackage{latexsym}

\usepackage[T1]{fontenc}

\usepackage[utf8]{inputenc}

\usepackage{microtype}

\usepackage{inconsolata}
\usepackage{graphicx}
\usepackage{multirow}
\usepackage{multicol}
\usepackage{color}
\usepackage{amsmath}
\usepackage{amsthm}
\usepackage{booktabs}
\usepackage{algorithm}
\usepackage{algorithmic}
\usepackage[switch]{lineno}
\title{DOP: Diagnostic-Oriented Prompting for Large Language Models in Mathematical Correction}

\author{Hao Chen,Biaojie Zeng, Xin Lin, Liang He \and Aimin Zhou \\
East China Normal University \\
\{10205102429,51265901131\}@stu.ecnu.edu.cn \\
\{xlin,lhe,amzhou\}@cs.ecnu.edu.cn
}

\begin{document}
\maketitle
\begin{abstract}
\textbf{Math world problems correction(MWPC)} is a novel task dedicated to rectifying reasoning errors in the process of solving mathematical problems. In this paper, leveraging the advancements in \textbf{large language models (LLMs)}, we address two key objectives:(1) Distinguishing between mathematical reasoning and error correction; (2) Exploring strategies to enhance the error correction capabilities of LLMs in mathematics to solve MWPC task. We noticed that, in real-time education,assisting students in recognizing their mistakes is more crucial than simply providing correct answers. However, current research tends to prioritize obtaining accurate solutions to math problems rather than correcting potentially incorrect ones. Therefore, we modify the research paradigm, demonstrating that improving mathematical reasoning abilities does not equate to mastery in error correction. Meanwhile, we propose a novel method called \textbf{diagnostic-oriented promping(DOP)} aimed at facilitating LLMs to excel in error correction. In experiments, DOP has shown outstanding performance, highlighting its significant impact. We argue that in mathematical education, the demand for outstanding correctors surpasses that for proficient reasoners. Codes and data are available on \href{https://github.com/ChenhaoEcnuCS/Reason-Correct}{\color{blue}{https://github.com/ChenhaoEcnuCS/Reason-Correct}}.
\end{abstract}

\section{Introduction}
\label{sec:intro}
\begin{quote}
    \textbf{``Give a man a fish and you feed him for a day; Teach a man to fish and you feed him for a lifetime.''}
    
    \hspace{3.8cm}  ----\textbf{Huainanzi}
\end{quote}

In recent years, the rapid advancement of large language models(LLMs)\cite{Zhao2023ASO:llmsurvey} has profoundly reshaped the landscape of artificial intelligence research.  The remarkable capabilities exhibited by prominent models like GPT-4 \cite{openai2023gpt4}, LLama2 \cite{llama2}, among others, have sparked innovative approaches across diverse domains of study.

In mathematics domain, numerous studies\cite{cot:start,zereshot-COT,Plan_and_Solve,self_consisit_COT,Auto_COT,Learn_From_Mistake,math_finetune,goat,metamath,wizardmath} have focused on the task of solving math world problems(MWPs). Some have employed diverse prompting strategies \cite{cot:start,zereshot-COT,Plan_and_Solve,self_consisit_COT,Auto_COT} to enhance the reasoning capabilities of LLMs, while others \cite{Learn_From_Mistake,math_finetune,goat,metamath,wizardmath} have fine-tuned models for mathematical tasks using domain-specific corpora.

However, we observe that most of these approaches primarily focus on achieving accuracy in solving MWPs. We often overlook the key point: merely enhancing the ability of a large language model to solve MWPs correctly falls short in mathematics pedagogy scenarios.

In real life, good students may be good at solving MWPs, but struggle to mentor their peers. Conversely, parents who may encounter difficulties in solving MWPs themselves can effectively coach their children using educational resources. This observation underscores the importance of \textbf{focusing not just on a model's ability to solve problems, but also on its capacity to correct errors and provide guidance.} With LLMs, an significant objective is instructing them to assist students in identifying and correcting their mistakes.

\begin{figure*}
  \centering
  \includegraphics[width=0.95\textwidth]{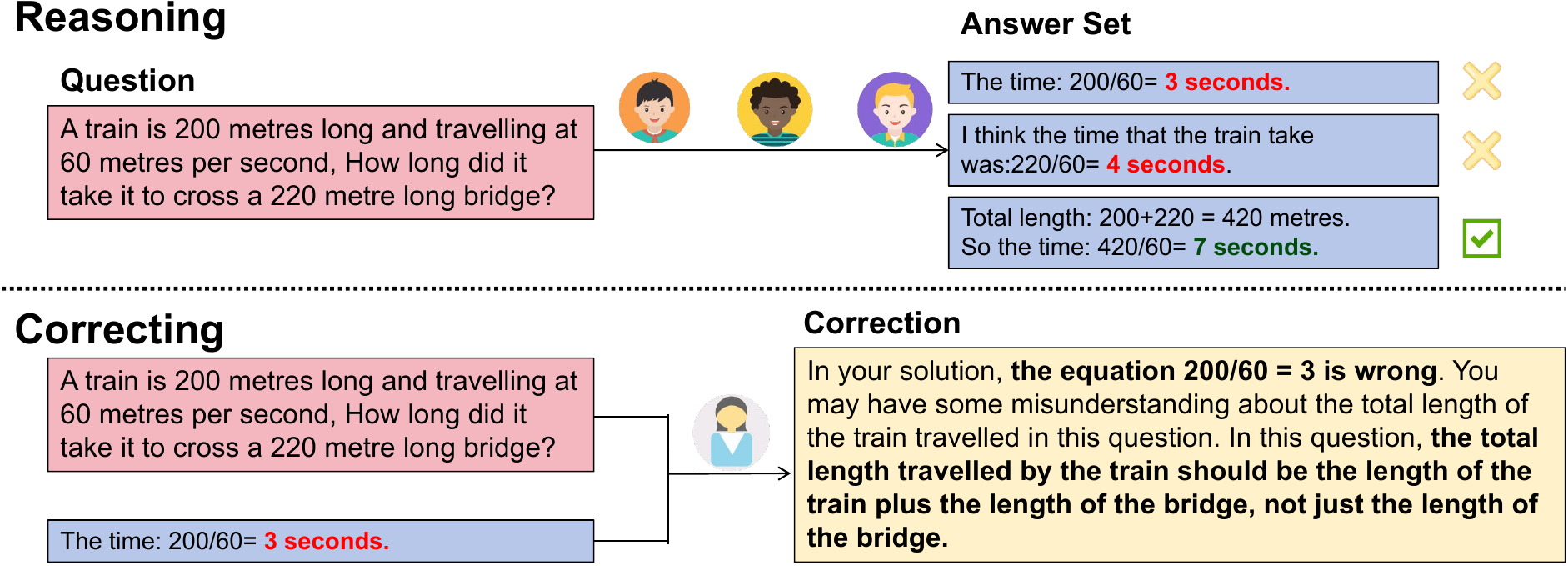}
  \caption{Examples of reasoning and correcting.}
  \label{fig:exampe_fig_1}
\end{figure*}

We first distinguish the concept of reasoning and correcting. As shown in Figure \ref{fig:exampe_fig_1}, in educational scenarios, the capacity for reasoning aids students in providing correct answers, whereas error correction empowers teachers to guide students through the process of identifying and rectifying mistakes in their responses. Our research mainly discussed those abilities in mathematics domain.

Therefore, we begin with a research question: \textbf{is the ability of a language model to reason and to correct errors equivalent?}

In some cases, an LLM may correctly solve a mathematical problem but fail to address errors in the solution. Conversely, it may inaccurately answer a math question but successfully rectify solution errors based on adequate contextual cues.

Based on the observation, we hypothesise that the reasoning and correcting capabilities are not fully equivalent. To demonstrate this, we introduce \textbf{math world problems correction(MWPC)}, a novel task focusing on the correction abilities of LLMs. We also conduct a series of experiments on MWPC task to prove our hypothesis, which will be described in Section 3.

Then, we further raise a question: \textbf{How can we enhance the correcting abilities of LLMs?}

\begin{figure*}
  \centering
  \includegraphics[width=0.95\textwidth]{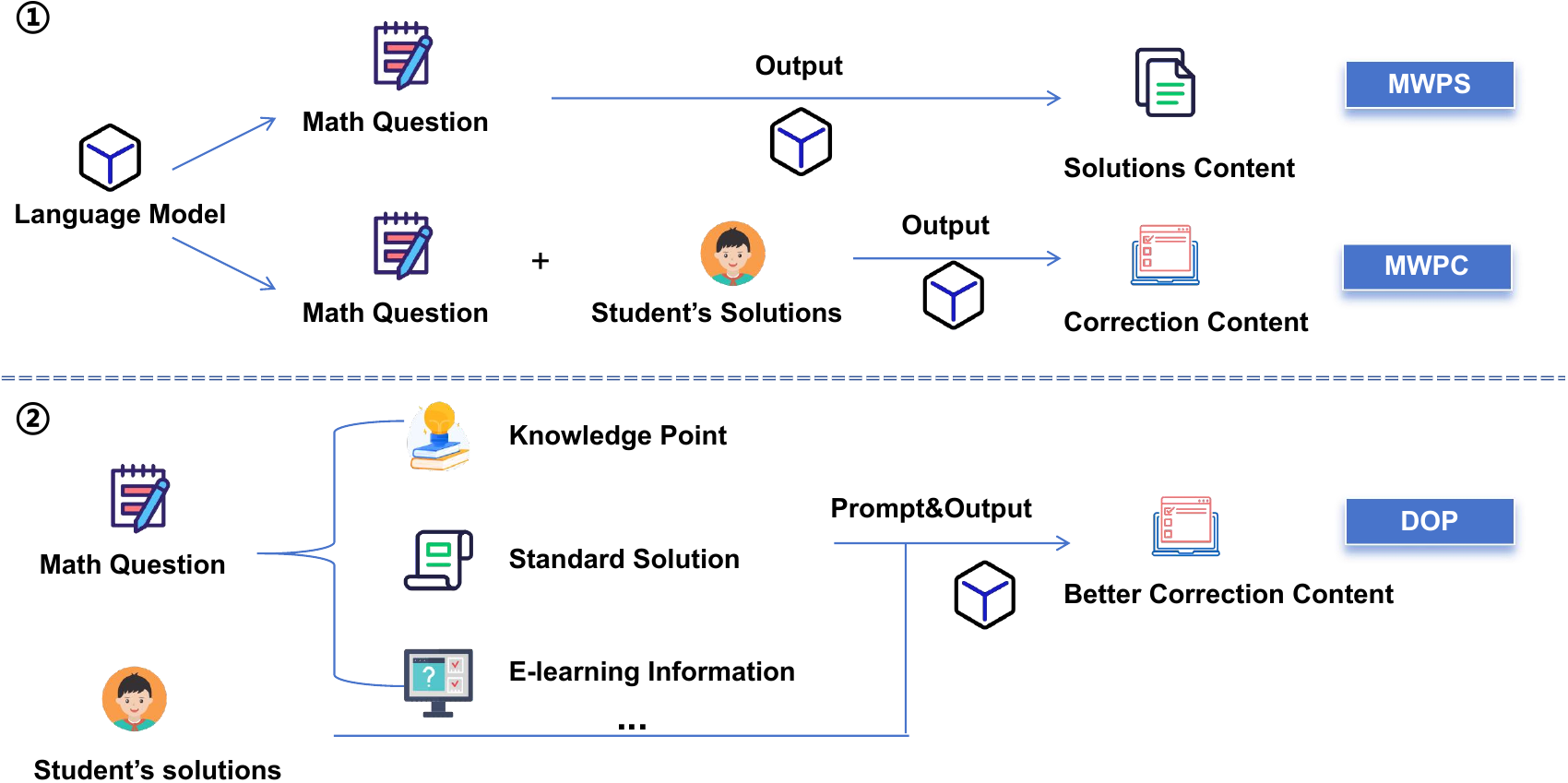}
  \caption{The overall framework of our research. In the first stage, we conduct both \textbf{MWPS} and \textbf{MWPC} tasks on our candidate models and prove that mathematical reasoning and correcting capabilities are not fully equivalent. Then, in the second stage, we conduct our strategy called \textbf{Diagnostic-Oriented Prompting(DOP)}, enabling our candidate models to enhance their correcting abilities in mathematical domain.}
  \label{fig:meth_3}
\end{figure*}

In modern teaching materials, both concise and detailed answers are commonly provided alongside the questions. Since we have demonstrated that the reasoning and correcting abilities were not fully equivalent, we proposed a novel method, called \textbf{Diagnostic-Oriented Prompting(DOP)}, leveraging available resources to enhance LLMs' proficiency as correctors in mathematical education.

Generally speaking, our contributions can be concluded as follows.

\begin{itemize}
    \item We modify the research paradigm, showing that in most LLMs, the abilities to reason and correct in MWPs are not fully equivalent, emphasizing that merely enhancing reasoning is insufficient.
    \item To the best of our knowledge, we are the first to propose MWPC task, which is more relevant and beneficial in mathematical education settings.
    \item We propose \textbf{Diagnostic-Oriented Prompting(DOP)}, a novel and effective method to enhance LLMs' correcting abilities based on modern teaching resources.
\end{itemize}

\section{Background and Related Work}

\subsection{Mathematical Reasoning Through LLMs}

There are many ways to improve the performance of LLMs on mathematical reasoning tasks by prompting them. 

The method of chain-of-thought(COT) prompting \cite{cot:start,zereshot-COT} can  be used in mathematical domain and improves the accuracy. \cite{self_consisit_COT} notices that a complex reasoning problem is usually thought of in a number of different ways and used majority voting to improve the process of COT. \cite{least_to_most,Plan_and_Solve} endeavour to decompose complex problems into multiple simple steps, guiding the large language model to solve mathematical problems step by step. \cite{X-of-thoughts,mathprompter,tora} mainly focus on using external tools like Python executor, mathematical calculator, and so on, to reduce the probability of error in LLMs and improve the reliability of LLMs in mathematical reasoning tasks.

In order to specifically enhance and utilise the mathematical reasoning ability of the model, some researchers use fine-tuning or instruction-tuning methods. \cite{reasoning_teachers} proposed fine-tuned COT, which generates reasoning samples from large teacher model to fine-tune smaller model. \cite{Learn_From_Mistake} utilised a corpus of mathematical reasoning containing error samples and the error correction process to fine-tune small models like LLama-2\cite{llama2} and MetaMath\cite{metamath}. \cite{goat} introduced Goat, which is a fine-tuned LLama model and can significantly outperforms GPT-4 \cite{openai2023gpt4} on a wide range of arithmetic tasks.

\subsection{Corrrection Throught LLMs}

Meanwhile, some research spotlights the correction capabilities of LLMs.

\cite{self-refine} proposed self-refine, which is a novel approach that allows LLMs to iteratively provide feedback and refine their own outputs. \cite{pan2023automatically} summarised a series of methods using feedback either produced by LLMs themselves or some external systems, to rectify those flaws. Self-correction effectively mitigates hallucination \cite{hallucination_survey} in LLMs. However, \cite{Cannot_self_correct} pointed out that without external feedback, LLMs still connot self-correct their own reasoning process, including mathematical reasoning process. According to \cite{gpt-4-wrong,llm-not-plan,llm-not-plan2,Cannot_self_correct}, when correcting something wrong, especially those errors produced by LLMs themselves, external information is indispensable.

There are also some studies centering on error correction task. \cite{remediation} used LLMs to remediate students' mathematical mistakes step by step. \cite{GEC_1,GEC_2,GEC_3,GEC_4} focused on grammatical error correction(GEC) task, utilising LLMs to solve GEC problems in monolingual and multilingual scenarios. \cite{BUG_1,BUG_2} researched the abilities of LLMs to correct errors in code, which is beneficial to computer science(CS) education. Unfortunately, there is still very little research on error correction to the mathematical reasoning process.

\subsection{AI For Mathematical Education}

Artificial Intelligence(AI) strongly promotes the development of mathematical education.

Since LLMs were put into use, \cite{remediation} simulated the process of human tutor, determining different strategy to address students' reasoning mistakes in mathematics. \cite{conic10k} studied mathematical education on conic sections in Chinese senior high school education using LLMs like GPT-4\cite{openai2023gpt4} and ChatGLM\cite{chatglm}. \cite{gpt-questioner} evaluated ChatGPT on generating pre-university math questions, providing insights for teachers and researchers in utilizing LLMs in mathematical education. The research above reveals that making LLMs to be good teachers is a following trend for AI in mathematical education.

\section{Methodology}

In this section, we will address the focus and describe the research methodology we used in this study.

Firstly, we conducted experiments to \textbf{validate differences between reasoning and correcting} in mathematics domain. Continue with the process, we proposed \textbf{Diagnostic-Oriented Prompting(DOP)} for correction capabilities. Figure \ref{fig:meth_3} shows the overall framework.

\subsection{Validating Differences between Reasoning and Correcting}

Initially, we conducted comparative experiments to validate the observation that the reasoning and error correction abilities of LLMs are not fully correlated.


We established several pivotal elements within this scenario. Firstly, our candidate models are represented as an expression $\boldsymbol{f(\cdot)}$, with the output sequence denoted as $\boldsymbol{y}$. In mathematical reasoning task, the input is a math question, denoted as $\boldsymbol{Q}$. We provided the model with a prompt containing the question, denoted as $\boldsymbol{P_r(Q)}$, and obtained an output $\boldsymbol{y_{r}}$, which means that:

\begin{align}
    \label{eq1}
    \boldsymbol{y_r=f(P_r(Q))}
\end{align}

Similarly, in the MWPC task, the input consists of a math question $\boldsymbol{Q}$ and its corresponding incorrect solution $\boldsymbol{W}$. We provided the model with a prompt containing both elements, denoted as $\boldsymbol{P_c(Q,W)}$, and obtained an output $\boldsymbol{y_c}$, indicating that:

\begin{align}
    \label{eq2}
    \boldsymbol{y_c=f(P_c(Q,W))}
\end{align}

In the next step, considering the question $\boldsymbol{Q}$, we examine standard answer, represented as $\boldsymbol{A}$. To ascertain the model's ability to solve the question, we employ an extraction function, denoted as $\boldsymbol{N_r}$ in the reasoning task and $\boldsymbol{N_c}$ in the correction task, to extract the final numeric answer from the natural language. Ultimately, we defined two states, $\boldsymbol{S_r}$ and $\boldsymbol{S_c}$, to indicate whether the model had successfully solved the task, which means that:

\begin{align}
    \label{eq3}
        \boldsymbol{S_r} & =
    \begin{cases}
      1, & \mbox{if}~~\boldsymbol{N_r(y_r)=N_r(A)}, \\
      0, & \mbox{otherwise}.
    \end{cases}\\
    \boldsymbol{S_c} & =
    \begin{cases}
      1, & \mbox{if}~~\boldsymbol{N_c(y_c)=N_c(A)}, \\
      0, & \mbox{otherwise}.
    \end{cases}
\end{align}

As mentioned above, it is necessary to collect a wide range of $\boldsymbol{\{Q,A,W\}}$ triplet. They are all represented as natural language.We chose several LLMs as our candidate models to perform both reasoning and correction tasks. Details about these selected models will be provided in Section \ref{sec:exp}.

\begin{figure*}
    \centering
    \includegraphics[width=0.98\textwidth]{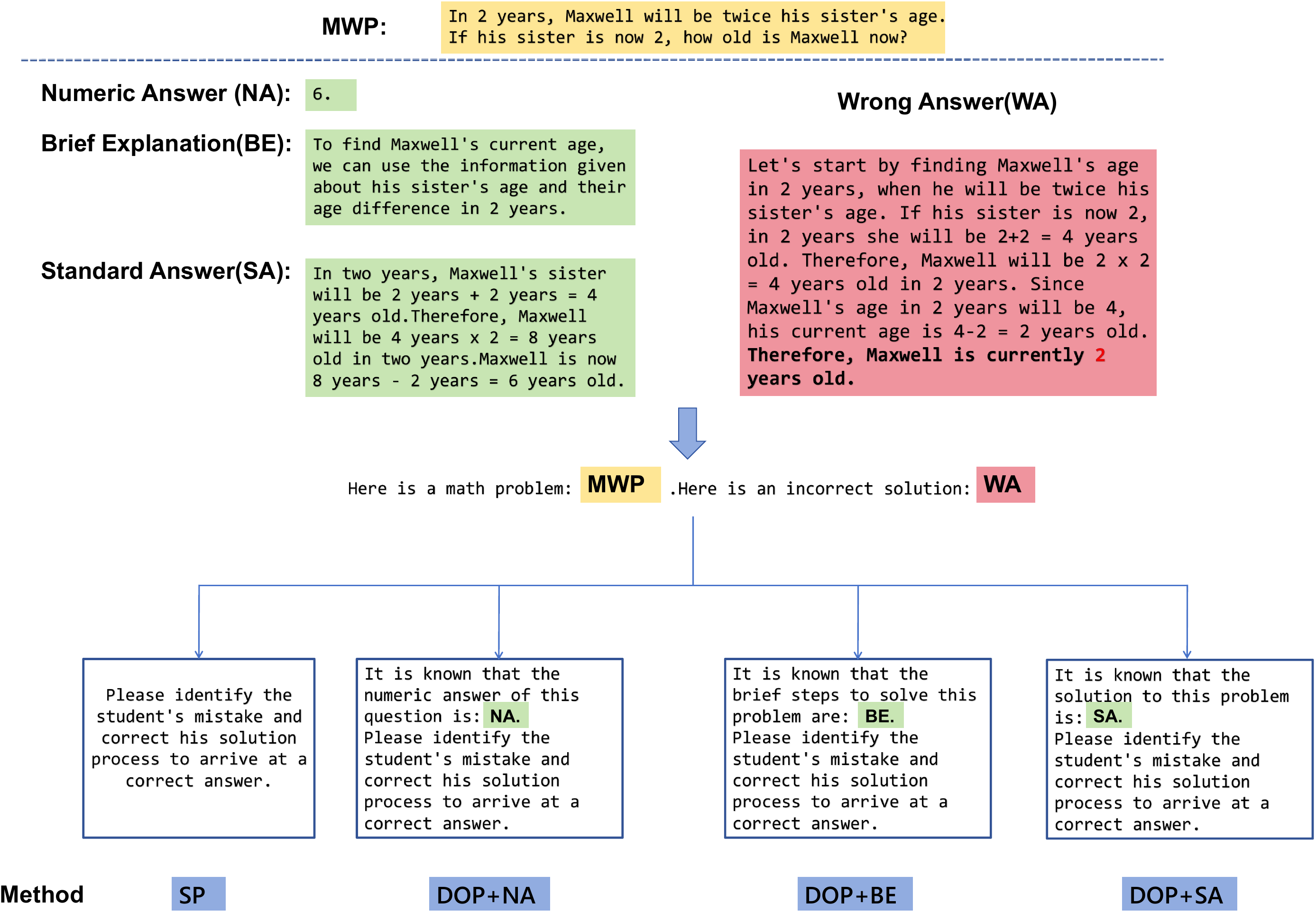}
    \caption{\centering An example of different levels of DOP.}
    \label{fig:DOP}
\end{figure*}  

\subsection{Diagnostic-Oriented Prompting(DOP)}

In our previous experiments, we observed that while LLMs may not entirely solve problems, they can generate correction processes. This parallels real-time education scenarios where teachers or parents, though unable to solve problems themselves, can guide children based on relevant information. Motivated by this, we propose a strategy named \textbf{Diagnostic-Oriented Prompting (DOP)} to leverage abundant resources and enhance the mathematical correction abilities of LLMs.

In modern educational materials, questions often come paired with answers, ranging from concise to detailed responses. Depending on the available resources, we can employ varying levels of DOP to enhance the correction abilities of LLMs. 

Furthermore, we conducted experiments involving 3 levels of DOP, affirming the effectiveness of the DOP approach.

The DOP framework comprises three levels, each with distinct input configurations. In the first level, the model's input consists of the mathematical problem, the erroneous solution and the correct numeric answer(NA) of the problem. In the second level, the model's input consists of the problem, the erroneous solution and the brief explanation(BE) of the problem. And finally, in the third level, the model's input consists of the problem, the erroneous solution and the standard answer(SA) of the problem. The prompt method that does not provide any additional supplementary information is labeled as standard prompting(SP).

The goal of DOP is to correct erroneous solution processes and arrive at the correct answer. The 3 levels of DOP progressively deepen and are denoted DOP+NA, DOP+BE and DOP+SA. The complete SP and DOP process is illustrated in Figure \ref{fig:DOP}.

\section{Expriments and Analysis}
\label{sec:exp}

\subsection{Experiment Setup}

We utilized some LLMs as candidate models, and collected multiple $\boldsymbol{\{Q,A,W\}}$ triplets from several mathematical datasets.

\textbf{Candidate models}.    We selected the following LLMs as out candidates, which contains some notable general models, some specialized mathematics models, and some educational-purpose models.

\begin{itemize}
    \item \textbf{GPT-4-0613\cite{openai2023gpt4}}.  GPT-4 is one of the most widely known LLMs, developed by openai. We selected the latest version.
    \item  \textbf{GPT-3.5-turbo\cite{openai2023gpt4}}.   A strong and remarkable model. It is also known as ChatGPT, developed by openai.
    \item  \textbf{LLama-2-Chat\cite{llama2}}.   LLama-2 is a collection of LLMs devloped by Meta and LLama-2-Chat is the fine-tuned model for dialogue use. We selected 3 parameter size: 7B, 13B and 70B.
    \item  \textbf{MetaMath\cite{metamath}}.   MetaMath is a fine-tuned model that specializes in mathematical reasoning.  Researchers used a rewrite strategy to bootstrap math questions and then fine tune the model. We selected 2 parameter size: 7B, 13B, pretrained from LLama2, and a 7B version pretrained on Mistral\cite{Jiang2023Mistral7}.
    \item  \textbf{WizardMath\cite{wizardmath}}.   WizardMath is a fine-tuned model using reinforcement learning from evol-instruct feedback for mathematical reasoning. We selected 2 parameter size: 7B, 13B.
    \item  \textbf{Baichuan2\cite{Baichuan}}.   Baichuan2 is a series of multilingual LLMs trained from scratch and perform well on some vertical domains including education. We selected 2 parameter size:7B, 13B.
\end{itemize}

\textbf{Data Construction}.    In our experiments, we collected sets of $\boldsymbol{{Q,A,W}}$ triplets, focusing on application problems in primary school mathematics described in natural language. The datasets we primarily referred to are as follows:

\begin{table*}
    \centering
    \begin{minipage}[t]{\textwidth}
    \centering
    \begin{tabular}{c|cc|cccc}
        \hline
        Model  & R-rate & C-rate & sR+sC & sR+uC & uR+sC & uR+uC \\
        \hline
        GPT-4-0613      & \textbf{0.859}  & \textbf{0.811}&  \textbf{2152}& 306 & 165 & 238 \\
        GPT-3.5-turbo   & 0.556  & 0.344&  659 & 932 & 325 & 945  \\
        LLama-2-chat-7b & 0.108  & 0.089 &  45  & 264 & 211 & 2341 \\
        LLama-2-chat-13b& 0.200  & 0.153&  148 & 424 & 290 & 1999    \\
        LLama-2-chat-70b& 0.318  & 0.224 & 282 & 629 & \textbf{358} & 1592 \\
        MetaMath-7b     & 0.764  & 0.180 &  455 & \textbf{1732}& 61  & 613  \\
        MetaMath-13b    & 0.772  & 0.238  & 606 & 1602& 76  & 577\\
        MetaMath-Mistral-7b & 0.733 & 0.254 & 637 & 1459 & 91 & 674\\
        WizardMath-7b   & 0.708  & 0.391 & 890 & 1138& 229 & 604 \\
        WizardMath-13b  & 0.486  & 0.165 & 294 & 1096& 177 & 1294 \\
        Baichuan-2-7b   & 0.079  & 0.059  & 29 & 196 & 139 & \textbf{2497}\\
        Baichuan-2-13b  & 0.281  & 0.105 & 133 & 690 & 186 & 1872\\
        \hline 
    \end{tabular}
    \caption{\centering The performance of candidate models in comparative experiments. The maximum value in each column is highlighted in \textbf{bold}.}
    \label{table:result_1}
    \end{minipage}
\end{table*}

\begin{itemize}
    \item \textbf{GSM8k\cite{GSM8k}}.    GSM8k is a  dataset of 8.5K high quality diverse grade school math word problems containing natural language solutions.
    \item \textbf{MathDial\cite{mathdial}}.   MathDial is a dataset of one-to-one teacher-student tutoring dialogues grounded in multi-step mathematical reasoning problems. Most of the math problems are from GSM8k.  
\end{itemize}

As MathDial provides problem statements, correct answers, and student confusion, we leveraged this data to construct a dataset comprising 2,861 sets of $\boldsymbol{\{Q,A,W\}}$ triplets.

\subsection{Results and Analysis} 

\begin{figure}
    \centering
    \includegraphics[width=0.47\textwidth]{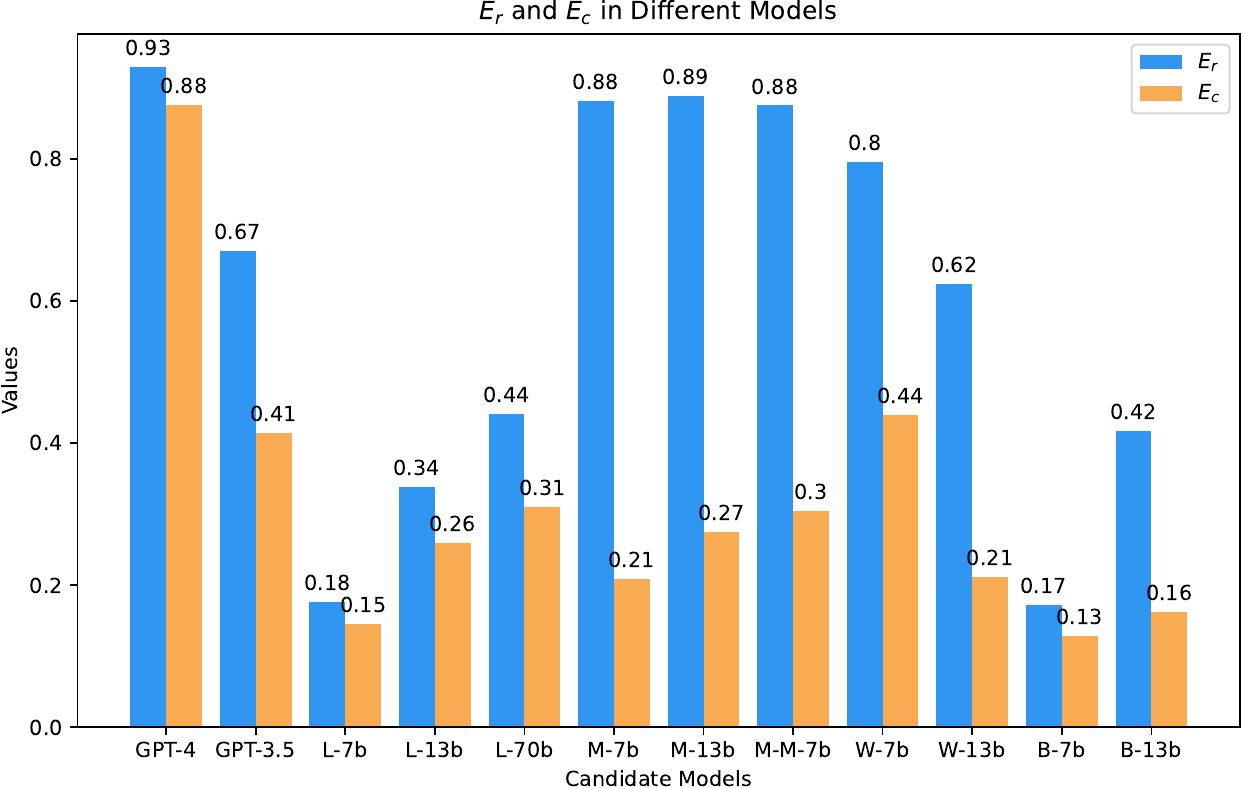}
    \caption{\centering Results of $E_r$ and $E_c$. We represents the candidate models using the first letters. For example, 'M-M-7b' means MetaMath-Mistral-7b.}
    \label{fig:E}
\end{figure}

\begin{table*}
    \begin{tabular}{p{3.8cm}|p{11.5cm}}
            \hline
            Question&Francine drives 140km to work each day. If she does not go to work 3 days every week, \textbf{find the total distance she drives to work for 4 weeks in kilometers.}\\\hline
            Ground-truth&There are 7 days in a week, so if he doesn't go to work for 3 days, he goes 7-3 = 4 days every week. He travels 140km each day for a weekly total of 140*4 = 560km. In 4 weeks he will travel a total of \textbf{\color[RGB]{1,142,29}{560 * 4 = 2240km.}} \\\hline
            Answer from GPT-4&Francine goes to work 7 - 3 = 4 days a week. This means that she drives a total of 140 * 4 =560km every week. Therefore in 4 weeks, she drives a total of \textbf{\color[RGB]{1,142,29}{560 * 4 = 2240km.}}\\\hline
            Student's solution&In a week, Francine drives \textbf{\color{red}{140km x 5 = 700km}} to work. In 4 weeks, she drives 700km x 4 = 2800km to work. But she does not go to work for 3 days every week, so in 4 weeks, she does not go to work for 3 x 4 = 12 days. Therefore, the total distance she drives to work in 4 weeks is \textbf{\color{red}{(700km x 4) - (140km x 12) = 2800km - 1680km = 1120km.}}\\\hline
            Correction From GPT-4&The student's mistake lies in the fact that they incorrectly subtracted the days she does not work from the total distance she drives to work. In reality, the days Francine does not work have already been accounted for when calculating the weekly distance. Here is the correct solution: In a week Francine drives 140km x 5 days = 700km to work.  In 4 weeks, she drives 700km x 4 = 2800km to work. There is no need to subtract the days she does not work because they have already been accounted for in the initial weekly calculation. \textbf{\color{red}{Answer: 2800km.}}\\\hline
        \end{tabular}
        \caption{\centering A case study of GPT-4. The error of this students lies in that he or she thinks there are five days in a week. GPT-4 effectively solved this problem. However, GPT-4 didn't capture the student's error.}
        \label{table:case_1}
\end{table*}

\subsubsection{Comparative Experiments for Validation.}

For each candidate model in the comparative experiments, we recorded the following information:

\begin{itemize}
    \item \textbf{R-rate}.\quad The rate of $\{Q,A,W\}$ triplets which were reasoned successfully.
    \item \textbf{C-rate}.\quad The rate of $\{Q,A,W\}$ triplets which were corrected successfully.
    \item \textbf{sR+sC}.\quad The number of $\{Q,A,W\}$ triplets which were reasoned and corrected successfully.
    \item \textbf{sR+uC}.\quad The number of $\{Q,A,W\}$ triplets which were reasoned successfully but corrected unsuccessfully.
    \item \textbf{uR+sC}.\quad The number of $\{Q,A,W\}$ triplets which were corrected successfully but reasoned unsuccessfully.
    \item \textbf{uR+uC}.\quad The number of $\{Q,A,W\}$ triplets which were reasoned and corrected unsuccessfully.
\end{itemize}

Table \ref{table:result_1} shows the performance of candidate models in comparative experiments. Let's start by analyzing the R-rate and C-rate. We can observe that GPT-4 achieves the highest performance both on MWPS and MWPC tasks. This indicates that as the most advanced general-purpose language model currently available, GPT-4's mathematical capabilities are clearly evident. Meanwhile, the specialized mathematics models like MetaMath show strong capabilities in mathematical reasoning, while their error correction abilities still have considerable room for improvement.

Next, we will analyze the performance of the models on the following four metrics: sR+sC, sR+uC, uR+sC and uR+uC. In Table \ref{table:result_1}, we can observe that, even in GPT-4, successfully solving a mathematical problem does not guarantee the ability to accurately correct an incorrect solution. Conversely, the model may not always provide an accurate solution, yet it can generate a proper correction process for an incorrect solution.

We further provides 2 definitions as follows.
\begin{align}
    \label{eq5}
        \boldsymbol{E_r} & =\frac{\boldsymbol{|sR+sC|}}{\boldsymbol{|(sR+sC)\cup(uR+sC)|}}
\end{align}

\begin{align}
    \label{eq6}
        \boldsymbol{E_c} & =\frac{\boldsymbol{|sR+sC|}}{\boldsymbol{|(sR+sC)\cup(sR+uC)|}}
\end{align}

As we mentioned above, $\boldsymbol{E_r}$ represents the ratio of the corrected numbers to the total reasoned numbers, while $\boldsymbol{{E_c}}$ represents the ratio of the reasoned numbers to the total corrected numbers. We displays the value of $\boldsymbol{E_r}$ and $\boldsymbol{E_c}$ in our experiments in Figure \ref{fig:E}.

In Figure \ref{fig:E}, we can observe that all our candidate models achieve higher $\boldsymbol{E_r}$ than $\boldsymbol{E_c}$. This suggests that if a model can successfully correct an error, it is more likely to solve the problem simultaneously. However, when the model is capable of solving a problem, the probability of correcting a related incorrect solution is much lower.

\begin{figure*}
    \centering
    \includegraphics[width=0.96\textwidth]{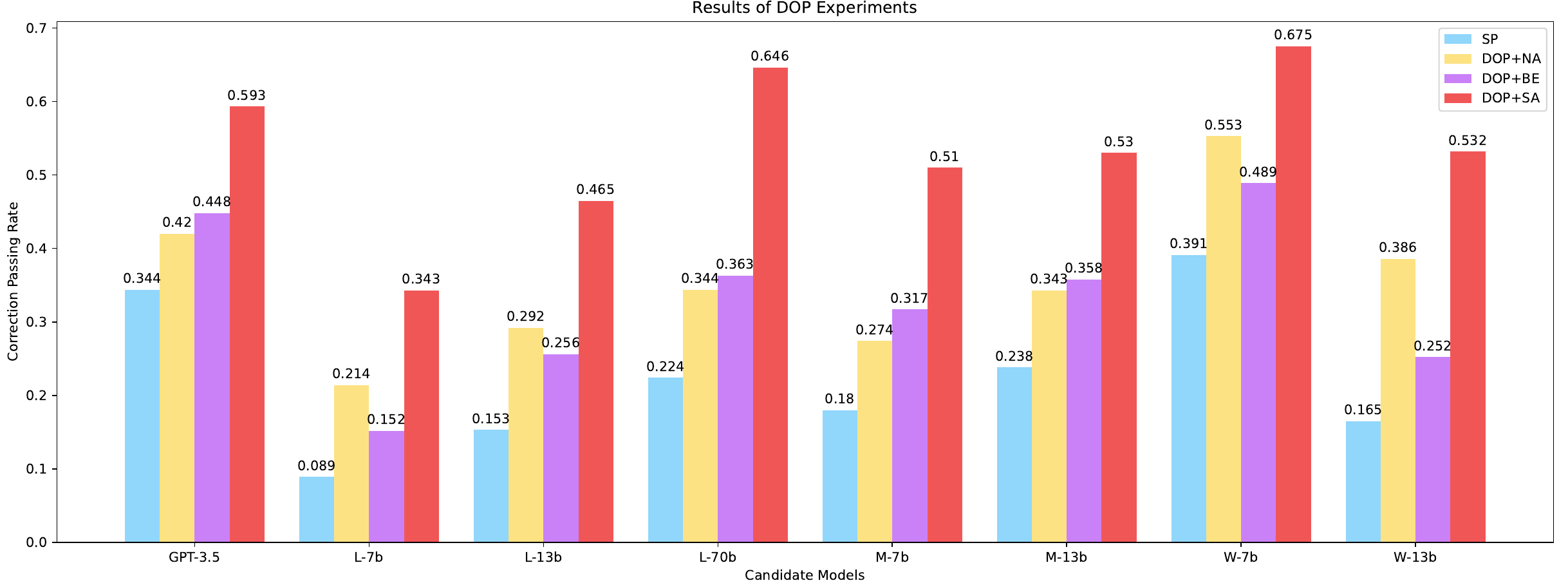}
    \caption{\centering Experiment results of DOP. We recorded the success rates of error correction under different scenarios and visualized them as bar charts.}
    \label{fig:DOP_result}
\end{figure*}

We also provide a case study in our experiment, as shown in Table \ref{table:case_1}. The mathematical problem requires finding the distance Francine has traveled during her 4-week work. GPT-4 effectively solved this problem. However, when faced with a student who miscalculated the number of working days, GPT-4 did not successfully correct its mistake. This indicates that for LLMs, successfully solving a mathematical problem does not necessarily mean they can successfully correct any errors that may arise within it. Similarly, successfully correcting an error within a mathematical problem does not imply that they can also successfully solve the problem.


To conclude, combining the result from Figure \ref{fig:E} and Table \ref{table:case_1}, we successfully demonstrate through comparative experiments that \textbf{the ability of LLMs in mathematical reasoning is not entirely equivalent to their ability in mathematical error correction}. Therefore, solely enhancing a model's mathematical problem-solving ability does not guarantee its proficiency as an error corrector. Further research is needed to thoroughly investigate the model's error correction capabilities in mathematics.

\subsubsection{Diagnostic-Oriented Prompting(DOP)}

For the DOP framework mentioned in Figure \ref{fig:meth_3} and Figure \ref{fig:DOP}, we conducted experiments involving 3 levels of DOP with several candidate models.




We studied DOP in 8 candidate models, comparing the correction passing rate between SP and DOP. We record the experimental results in Figure \ref{fig:DOP_result}.

We found that when employing DOP, all candidate models achieved higher pass rates compared to using SP alone during the MPWC task. This suggests that the DOP method significantly enhances the mathematical error-correction capabilities of LLMs.

\section{Conclusions}

In this paper, we have come to the following conclusions.

\textbf{1.LLMs' reasoning and correcting abilities are not fully equivalent.} In our comparative experiment, LLMs may solve a problem but fail to correct a wrong solution of this problem. Also, they may not solve a problem properly, but can find reasoning errors and correct them in a wrong solution. 

\textbf{2.Mainstream LLMs' have stronger reasoning abilities than correcting abilities.} In our experiments, our candidate models perform better in reasoning task than correcting tasks. This suggests that while LLMs excel as reasoners, their ability to correct errors is limited. Therefore, further research into their correction abilities is necessary. 

\textbf{3.Improving LLMs' correcting abilities is vital and essential.} In mathematical education scenarios, it is more vital to correct the error from the students, rather than merely providing solutions. Since we have demonstrated that reasoning and correcting abilities are not the same thins, and reasoning abilities are much better, improving LLMs' correcting abilities bocomes vital and important.

\textbf{4.Diagnostic-Oriented Prompting(DOP) is an effective method to enhance the correcting abilities of LLMs.} We modify the research paradigm of the mainstream research and proposes MWPC task. With the aid of educational resources and DOP, LLMs can be an excellent corrector, which is useful to help students dealing with understanding math world problems.

\section{Limitations and Future Work}

We have several limitations in this work.Firstly, there are still lack of high-quality mathematical correction datasets to study the relative abilities of LLMs. Meanwhile, we study correction mainly based on all kinds of language models. In fact, the behaviour of LLMs and human teachers and students differs a lot. We still need deeper research in the field. To study this issue well, our future work is as follows:

\begin{itemize}
    \item \textbf{Collect high-quality MWPs and corresponding mistakes.} High-quality data is vital for us to enhance the performance of LLMs. Most mainstream datasets in mathematical domain are lack of some relevant solutions with errors, which is not helpful to study the correcting abilities of LLMs. As a result, we are committed to construct a high-quality dataset containg MWPs and corresponding mistakes.
    \item \textbf{We need a deeper view of real-time mathematical education scenarios.} The behaviours between human and language models differs a lot. We also need some data from the real life, not just merely from the language models. In the future, it is necessary for us to go deeper to the real-time education scenarios.
    \item \textbf{Develop more level of DOP.} We have broken the mold and proven the effectiveness of DOP. It is still necessary to develop a higher level of DOP method.

\end{itemize}

\bibliography{custom}

\appendix



\end{document}